%% file: main.tex
\documentclass[10pt,twocolumn,letterpaper]{article}

\usepackage{wacv}              %

\input{preamble}

\definecolor{wacvblue}{rgb}{0.21,0.49,0.74}
\usepackage[pagebackref,breaklinks,colorlinks,allcolors=wacvblue]{hyperref}

\def\wacvPaperID{2957} %
\def\confName{WACV}
\def\confYear{2027}

\title{HyDRA: Hybrid Domain-Aware Robust Architecture \\ for Heterogeneous Collaborative Perception}

\author{Minwoo Song \qquad Minhee Kang \qquad Heejin Ahn\\
Korea Advanced Institute of Science and Technology (KAIST), Republic of Korea\\
{\tt\small \{haestle1, ministop, heejin.ahn\}@kaist.ac.kr}}

\begin{document}
\maketitle
{\let\thefootnote\relax\footnotetext{\copyright~2027 IEEE. Personal use of this material is permitted. Permission from IEEE must be obtained for all other uses, in any current or future media, including reprinting/republishing this material for advertising or promotional purposes, creating new collective works, for resale or redistribution to servers or lists, or reuse of any copyrighted component of this work in other works.}}
\input{sec/0_abstract}    
\input{sec/1_intro}
\input{sec/2_problem}
\input{sec/3_hydra}
\input{sec/4_experiments}
\input{sec/5_conclusion}

\noindent\textbf{Acknowledgments.} This work was partly supported by the Institute of Information \& Communications Technology Planning \& Evaluation (IITP)-ITRC (Information Technology Research Center) grant funded by the Korea government (MSIT) (IITP-2027-RS-2023-00259991, 50\%) and the National Research Foundation of Korea (NRF) grant funded by the Ministry of Science and ICT (MSIT) (RS-2025-16072587, 50\%).

{
    \small
    \bibliographystyle{ieeenat_fullname}
    \bibliography{main}
}

\clearpage
\setcounter{figure}{0}\setcounter{table}{0}\setcounter{equation}{0}
\renewcommand{\thefigure}{A\arabic{figure}}
\renewcommand{\thetable}{A\arabic{table}}
\renewcommand{\theequation}{A\arabic{equation}}

\maketitlesupplementary   %

\appendix

\section{Implementation Details}
\label{sec:supp_impl}

We report the hyperparameters omitted from the main paper for brevity. All detector backbones follow the training configuration described in the main paper (AdamW, batch size $2$, $25$ epochs); HyDRA's three modules introduce no additional training and operate entirely at inference time.

\paragraph{Domain classifier.}
In Step~1, predictions and pseudo-ground truths are matched with the Hungarian algorithm, and a matched pair is retained only if its IoU exceeds $0.2$.
In Step~2, the temperature controlling confidence sensitivity is set to $\sigma = 1.0$.

\paragraph{Anchor-guided pose graph optimization (AG-PGO).}
An edge between a heterogeneous detection and an anchor is established when their center distance is below $3.0$\,m and their yaw difference is below $20^{\circ}$.
The information-matrix exponents are set to $\gamma = 2.0$ and $\beta = 1.5$.
The pose graph is optimized with g2o for up to $50$ iterations.

\paragraph{PHCP evaluation protocol.} PHCP's official implementation supports OPV2V/OPV2V-H only and evaluates via scenario-level few-shot splits with a per-scenario adapter fine-tuning procedure, distinct from the fixed train/test protocol used by the other baselines in this paper. V2X-Real is not supported. We therefore report only its adaptation latency, as reported in the original work.

\section{Domain Classifier Score Analysis}
\label{sec:supp_domain_score}

\begin{table}[t]
\centering
\setlength{\tabcolsep}{8pt}
\resizebox{\columnwidth}{!}{%
\begin{tabular}{cc ccc}
\toprule
\multirow{2}{*}{\textbf{Agent Type}} & \multirow{2}{*}{\textbf{Noise Setup}} & \multicolumn{3}{c}{\textbf{Domain Score ($S_{\text{domain}}$)}} \\
\cmidrule(lr){3-5}
 &  & \textbf{Mean} & \textbf{Max} & \textbf{Min} \\
\midrule
\multirow{2}{*}{$H_P$ (Hom.)} & w/o Noise & 0.5136 & 0.7046 & 0.2458 \\
 & w/ Noise & 0.5135 & 0.7013 & 0.2361 \\
\midrule
\multirow{2}{*}{$X_P$ (Het.)} & w/o Noise & 0.0072 & 0.1444 & 0.0000 \\
 & w/ Noise & 0.0070 & 0.1421 & 0.0000 \\
\midrule
$X_P$ (seed 2) & w/o Noise & 0.0018 & 0.0423 & 0.0000 \\
$X_P$ (seed 3) & w/o Noise & 0.0090 & 0.1660 & 0.0000 \\
\midrule
\multirow{2}{*}{$X_S$ (Het.)} & w/o Noise & 0.0001 & 0.0134 & 0.0000 \\
 & w/ Noise & 0.0001 & 0.0101 & 0.0000 \\
\midrule
\multirow{2}{*}{$X_V$ (Het.)} & w/o Noise & 0.0001 & 0.0043 & 0.0000 \\
 & w/ Noise & 0.0001 & 0.0027 & 0.0000 \\
\bottomrule
\end{tabular}%
}
\caption{Analysis of domain classifier scores $S_{\text{domain}}$ for homogeneous and heterogeneous agents, with and without pose noise. Rows labeled ``seed 2/3'' report $X_P$ instances retrained with different random seeds.}
\label{tab:domain_score}
\end{table}

\Cref{tab:domain_score} reports the full per-type breakdown of the domain score $S_{\text{domain}}$ produced by our domain classifier, under both noise-free and noisy pose conditions.
Homogeneous agents ($H_P$) consistently yield high scores (mean $\approx 0.51$), while all heterogeneous agents ($X_P$, $X_S$, $X_V$) collapse toward zero (mean $\le 0.01$).
The two groups remain fully separated even in the worst case: the minimum homogeneous score ($0.2361$) still exceeds the highest heterogeneous score observed across all instances, including the two additional $X_P$ seeds ($0.1660$), leaving a margin of roughly $0.07$.
Any routing threshold within this gap cleanly separates the two regimes, confirming that the classifier does not require precise threshold tuning.

The scores are also robust to pose noise: across all agent types, adding noise shifts the mean domain score by at most $0.0002$, preserving the separation margin.
Since $X_P$ corresponds to a single independently trained checkpoint, we verify that the observed separation is not an artifact of that particular training run by retraining $X_P$ with two additional random seeds under an identical recipe.
Both additional checkpoints yield domain scores well below the minimum $H_P$ score (max $0.0423$ and $0.1660$, respectively, versus $H_P$'s minimum of $0.2361$--$0.2458$), 
confirming that the classifier separates $X_P$ from $H_P$ consistently across independently trained instances. Both additional checkpoints follow the same single-agent recipe as $X_P$; they test robustness to the training run, not to the training regime.
These results support routing homogeneous agents to intermediate fusion and heterogeneous agents to late fusion using $S_{\text{domain}}$ alone.

\paragraph{Generalization to OPV2V.}
\Cref{tab:domain_score_opv2v} reports the same score breakdown on OPV2V.
The separation observed on V2X-Real transfers directly: the homogeneous agent ($H_P$) scores highly (mean $0.5661$), while all heterogeneous agents collapse toward zero (mean $\le 0.0007$), and the minimum homogeneous score ($0.4419$) still clears the maximum heterogeneous score across all three heterogeneity types ($0.0308$, from $X_P$), yielding a feasible threshold region of $[0.031, 0.442]$.
This is consistent with the region reported for V2X-Real and confirms that a single threshold requires no per-dataset recalibration, corroborating the claim in the main paper that this separation generalizes across datasets.

\begin{table}[t]
\centering
\setlength{\tabcolsep}{8pt}
\begin{tabular}{c ccc}
\toprule
\multirow{2}{*}{\textbf{Agent Type}} & \multicolumn{3}{c}{\textbf{Domain Score ($S_{\text{domain}}$)}} \\
\cmidrule(lr){2-4}
 & \textbf{Mean} & \textbf{Max} & \textbf{Min} \\
\midrule
$H_P$ (Hom.) & 0.5661 & 0.7004 & 0.4419 \\
\midrule
$X_P$ (Het., latent) & 0.0007 & 0.0308 & 0.0000 \\
$X_S$ (Het., arch.)  & 0.0002 & 0.0084 & 0.0000 \\
$X_V$ (Het., arch.)  & 0.0000 & 0.0000 & 0.0000 \\
\bottomrule
\end{tabular}
\caption{Domain classifier scores $S_{\text{domain}}$ on OPV2V, under the identical agent configuration used for V2X-Real in \cref{tab:domain_score}.}
\label{tab:domain_score_opv2v}
\end{table}

\paragraph{Mean-IoU ablation.}
We further check that this separation is not simply an artifact of using IoU as the matching cost, by replacing Soft-AP with a purely geometric mean-IoU score---the mean IoU between each matched pseudo-ground truth and its locally decoded detection, without the confidence-consistency term or the precision--recall integration. Under the same full-observation setting used above, this ablation collapses the separation entirely: homogeneous and heterogeneous score distributions overlap substantially, and no feasible threshold region can be identified. This confirms that Soft-AP's separation stems from its confidence-aware, rank-integrated formulation rather than from IoU matching alone.

\section{Generality Across Ego Architectures}
\label{sec:supp_ego_generality}

\begin{table}[t]
\centering
\setlength{\tabcolsep}{8pt}
\resizebox{\columnwidth}{!}{%
\begin{tabular}{cl ccc}
\toprule
\multirow{2}{*}{\textbf{Ego}} & \multirow{2}{*}{\textbf{Agent Type}} & \multicolumn{3}{c}{\textbf{Domain Score ($S_{\text{domain}}$)}} \\
\cmidrule(lr){3-5}
 & & \textbf{Mean} & \textbf{Max} & \textbf{Min} \\
\midrule
\multirow{4}{*}{$E_S$}
 & $H_S$ (Hom.) & 0.5747 & 0.7415 & 0.2303 \\
 & $X_S$ (Het., latent) & 0.0000 & 0.0193 & 0.0000 \\
 & $X_P$ (Het., arch.) & 0.0000 & 0.0000 & 0.0000 \\
 & $X_V$ (Het., arch.) & 0.0021 & 0.0698 & 0.0000 \\
\bottomrule
\end{tabular}%
}
\caption{Domain classifier scores $S_{\text{domain}}$ when the ego agent uses SECOND ($E_S$) instead of PointPillars. ``Het., latent'' denotes an auxiliary sharing the ego's architecture but trained independently (analogous to $X_P$ under $E_P$); ``Het., arch.'' denotes an auxiliary using one of the two remaining backbones.}
\label{tab:supp_ego_generality}
\end{table}

All experiments in the main paper use PointPillars as the ego backbone ($E_P$). To verify that our domain classifier is not specific to this choice, we repeat the score analysis of \cref{sec:supp_domain_score} with the ego agent instantiated as SECOND ($E_S$).

\Cref{tab:supp_ego_generality} reports the results. For $E_S$, the homogeneous agent again yields a high domain score (mean $0.57$) while every heterogeneous agent---whether latent-domain or architecturally distinct---collapses toward zero (mean $\le 0.0021$). The minimum homogeneous score exceeds the maximum heterogeneous score by a margin of $0.16$ ($0.2303$ vs.\ $0.0698$), an even wider separation than the $0.07$ margin observed for $E_P$ in \cref{sec:supp_domain_score}. These results confirm that the classifier's separation between homogeneous and heterogeneous agents does not depend on a particular choice of ego backbone.

\section{Domain Classifier Latency}
\label{sec:supp_latency}

\begin{table}[t]
\centering
\setlength{\tabcolsep}{8pt}
\begin{tabular}{lcc}
\toprule
\multirow{2}{*}{\textbf{Stage}} & \multicolumn{2}{c}{\textbf{Latency (ms)}} \\
\cmidrule(lr){2-3}
 & \textbf{Mean} & \textbf{Std} \\
\midrule
Decode          & $6.650$ & $0.290$ \\
Post-processing & $0.723$ & $0.093$ \\
Match           & $2.053$ & $2.283$ \\
\midrule
\textbf{Total}  & $\mathbf{9.426}$ & $\mathbf{2.334}$ \\
\bottomrule
\end{tabular}
\caption{Per-agent, per-frame latency breakdown of the domain classifier (2{,}553 measurements).}
\label{tab:classifier_latency}
\end{table}

\Cref{tab:classifier_latency} breaks down the inference latency of the domain classifier into its three stages: feature decoding, post-processing, and object matching.
Latency is measured per auxiliary agent per frame over the evaluation set, yielding 2{,}553 measurements.
All measurements are taken on the same single NVIDIA RTX 4090 GPU used for training.

The decoding stage dominates the cost ($6.65$ ms) but is highly stable ($\pm 0.29$ ms), as it processes a feature map independent of scene content.
Post-processing is negligible ($0.72$ ms).
The matching stage is the only data-dependent component: its latency scales with the number of matched detection pairs $|\mathcal{M}|$ between the ego-decoded predictions $\mathcal{B}_{\text{pred}}$ and the auxiliary detections $\mathcal{B}_A$. This makes its cost markedly higher for homogeneous agents, whose features the ego decoder interprets well and thus yield many matched pairs, and lower for heterogeneous agents, where few predictions survive matching---which also explains the large variance of this stage ($\pm 2.28$ ms).

\section{Subsampling Stress-Test Analysis}
\label{sec:supp_subsampling}

We detail the protocol behind the sparse-observation stress-test reported in the main paper, and provide the distribution spread that explains the reported error trends.

\paragraph{Protocol.}
For a target count $K$, we randomly subsample $\min(K, S)$ of the $S$ shared objects of each auxiliary agent without replacement, and restrict the ego detections to those lying within $3.0$\,m of any subsampled auxiliary object before computing the domain score $S_{\text{domain}}$.
To average out sampling variance, we repeat this procedure $R=30$ times per frame and aggregate the resulting scores over all frames with at least one shared object.
The misclassification rate (Err.) is then measured against the routing threshold $\tau=0.135$: a homogeneous agent is counted as misclassified when $S_{\text{domain}} < \tau$, and a heterogeneous agent when $S_{\text{domain}} \geq \tau$.
This isolates the effect of observation sparsity on classification reliability, independent of the specific objects sampled.

\begin{table}[t]
\centering
\small
\setlength{\tabcolsep}{4pt}
\begin{tabular}{c l ccc c}
\toprule
\textbf{$K$} & \textbf{Scoring} & \textbf{AUROC} & \textbf{Hom.\ (\%)} & \textbf{Het.\ (\%)} & \textbf{Sum (\%)} \\
\midrule
\multirow{2}{*}{1}  & Soft-AP & 0.9633 & 3.09 & 3.63 & 6.72 \\
                     & Hard-AP & 0.9594 & 3.09 & 3.63 & 6.72 \\
\midrule
\multirow{2}{*}{2}  & Soft-AP & 0.9679 & 0.34 & 6.99 & 7.33 \\
                     & Hard-AP & 0.9629 & 0.19 & 7.16 & 7.35 \\
\midrule
\multirow{2}{*}{3}  & Soft-AP & 0.9725 & 0.76 & 6.88 & 7.64 \\
                     & Hard-AP & 0.9589 & 0.86 & 6.54 & 7.40 \\
\midrule
\multirow{2}{*}{5}  & Soft-AP & 0.9878 & 1.69 & 6.69 & 8.38 \\
                     & Hard-AP & 0.9766 & 0.10 & 10.56 & 10.66 \\
\midrule
\multirow{2}{*}{10} & Soft-AP & 0.9985 & 0.76 & 1.54 & 2.30 \\
                     & Hard-AP & 0.9972 & 0.45 & 3.22 & 3.67 \\
\midrule
\multirow{2}{*}{20} & Soft-AP & 1.0000 & 0.03 & 0.20 & 0.23 \\
                     & Hard-AP & 1.0000 & 0.04 & 0.28 & 0.32 \\
\bottomrule
\end{tabular}
\caption{Soft-AP vs.\ Hard-AP under the single-frame ($N=1$) subsampling protocol ($n=77{,}190$ pooled samples per $K$). For each scoring function and $K$, the threshold is independently searched to minimize $\text{hom\_misclass}+\text{het\_misclass}$ (not the fixed $\tau=0.135$ used elsewhere in this stress test). Sum is this minimized objective, $\text{hom\_misclass}+\text{het\_misclass}$.}
\label{tab:supp_softap}
\end{table}

\paragraph{Soft-AP vs.\ Hard-AP.}
Recall that Soft-AP's per-detection quality score is $q_k$, combining a continuous IoU term with a confidence-consistency term between the ego and auxiliary detections. We validate this formulation against a Hard-AP variant that replaces $q_k$ with a binary indicator---$1$ if the matched pair's IoU is at least $0.3$ and $0$ otherwise---thereby removing both the continuous IoU weighting and the confidence-consistency term at once, while keeping the Hungarian matching and the rank-based AP integration unchanged. This binary criterion is stricter than the IoU $>0.2$ gate used to retain Hungarian matches (\cref{sec:supp_impl}), so it discards a subset of the retained pairs rather than acting as a no-op; it is fixed at $0.3$ and not swept. Because this ablation collapses two design choices into one comparison, it isolates their joint contribution rather than either alone. Under the same single-frame ($N=1$) subsampling protocol above, we search, independently for each scoring function and each $K$, the score threshold that minimizes $\text{hom\_misclass} + \text{het\_misclass}$ over the pooled $n=77{,}190$ samples---distinct from the fixed stress-test operating threshold $\tau=0.135$ used elsewhere in this stress test. \Cref{tab:supp_softap} reports the resulting AUROC and misclassification rates.

Across the threshold-independent AUROC metric, Soft-AP matches or exceeds Hard-AP at every $K$, with the largest gap at $K=3$ despite the threshold-optimized sum favoring Hard-AP there.
$K\le 3$ remains near-degenerate for both variants; neither scoring function is reliable enough to route on directly at these sparsity levels. Once $K\ge 5$, Soft-AP consistently yields a lower Hom.+Het. sum than Hard-AP, confirming that the continuous, confidence-aware formulation of Soft-AP is the better choice for our routing decision.

\begin{table}[t]
\centering
\setlength{\tabcolsep}{6pt}
\resizebox{\columnwidth}{!}{%
\begin{tabular}{cl cccccc}
\toprule
\multirow{2}{*}{\textbf{Agent}} & \multirow{2}{*}{\textbf{Metric}} & \multicolumn{6}{c}{\textbf{Shared Objects per Frame ($K$)}} \\
\cmidrule(lr){3-8}
 & & \textbf{1} & \textbf{2} & \textbf{3} & \textbf{5} & \textbf{10} & \textbf{20} \\
\midrule
\multirow{2}{*}{$H_P$}
 & $S_{\text{domain}}$ & 0.288 & 0.251 & 0.241 & 0.239 & 0.267 & 0.350 \\
 & Err. (\%) & 41.94 & 29.58 & 22.74 & 13.14 & 2.31 & 0.03 \\
\midrule
\multirow{2}{*}{$X_P$}
 & $S_{\text{domain}}$ & 0.014 & 0.014 & 0.014 & 0.014 & 0.013 & 0.012 \\
 & Err. (\%) & 2.99 & 4.66 & 5.27 & 2.87 & 0.96 & 0.22 \\
\midrule
\multirow{2}{*}{$X_{S/V}$}
 & $S_{\text{domain}}$ & 0.000 & 0.000 & 0.000 & 0.000 & 0.000 & 0.000 \\
 & Err. (\%) & 0.07 & 0.10 & 0.09 & 0.02 & 0.00 & 0.00 \\
\bottomrule
\end{tabular}%
}
\caption{Mean domain score $S_{\text{domain}}$ and misclassification rate (Err.) at $N=1$ under subsampled shared observations ($K$), at threshold $\tau=0.135$.}
\label{tab:supp_subsampling_mean}
\end{table}

\begin{table}[t]
\centering
\setlength{\tabcolsep}{5pt}
\resizebox{\columnwidth}{!}{%
\begin{tabular}{cl cccccc}
\toprule
\multirow{2}{*}{\textbf{Agent}} & \multirow{2}{*}{\textbf{Pct.}} & \multicolumn{6}{c}{\textbf{Shared Objects per Frame ($K$)}} \\
\cmidrule(lr){3-8}
 & & \textbf{1} & \textbf{2} & \textbf{3} & \textbf{5} & \textbf{10} & \textbf{20} \\
\midrule
\multirow{2}{*}{$H_P$}
 & p1  & 0.000 & 0.041 & 0.062 & 0.084 & 0.122 & 0.183 \\
 & p99 & 0.898 & 0.768 & 0.670 & 0.563 & 0.515 & 0.580 \\
\midrule
\multirow{2}{*}{$X_P$}
 & p1  & 0.000 & 0.000 & 0.000 & 0.000 & 0.000 & 0.000 \\
 & p99 & 0.550 & 0.337 & 0.246 & 0.183 & 0.134 & 0.098 \\
\bottomrule
\end{tabular}%
}
\caption{Distribution spread (1st/99th percentiles) of the domain score $S_{\text{domain}}$ under subsampling. The threshold is $\tau=0.135$.}
\label{tab:subsampling_spread}
\end{table}

\paragraph{Mean domain score under sparsity.}
\Cref{tab:supp_subsampling_mean} reports the mean domain score $S_{\text{domain}}$ alongside the misclassification rate (Err.) at $N=1$ (single-frame, no accumulation), complementing the percentile spread in \cref{tab:subsampling_spread}. For $H_P$, the mean score dips at intermediate $K$ ($0.239$--$0.251$ for $K=2,3,5$) before recovering at $K=20$ ($0.350$), reflecting that a small, unluckily sampled subset of shared objects can pull the average down even though the lower tail (\cref{tab:subsampling_spread}) is what ultimately drives misclassification. For $X_P$, the mean stays essentially flat around $0.011$--$0.015$ regardless of $K$, well below the routing threshold $\tau=0.135$ used in this stress test.

\begin{table*}[t]
\centering
\resizebox{\textwidth}{!}{%
\begin{tabular}{l cccc cccc cccc}
\toprule
\multirow{2}{*}{\textbf{Method}} & \multicolumn{4}{c}{\textbf{AP@0.3}} & \multicolumn{4}{c}{\textbf{AP@0.5}} & \multicolumn{4}{c}{\textbf{AP@0.7}} \\
\cmidrule(lr){2-5}\cmidrule(lr){6-9}\cmidrule(lr){10-13}
 & \textbf{vehicle} & \textbf{pedestrian} & \textbf{truck} & \textbf{total} & \textbf{vehicle} & \textbf{pedestrian} & \textbf{truck} & \textbf{total} & \textbf{vehicle} & \textbf{pedestrian} & \textbf{truck} & \textbf{total} \\
\midrule
Late Fusion & 0.5244 & \textbf{0.2390} & 0.3719 & 0.3784 & 0.3568 & \textbf{0.1084} & 0.1911 & 0.2188 & 0.0691 & \textbf{0.0046} & 0.0685 & 0.0474 \\
E2E Training & \underline{0.8390} & 0.0951 & 0.5213 & 0.4851 & 0.7257 & 0.0153 & 0.4597 & 0.4002 & 0.2580 & 0.0001 & 0.2250 & 0.1610 \\
MPDA & 0.8096 & 0.1536 & \textbf{0.5412} & \underline{0.5014} & \textbf{0.7533} & 0.0306 & \textbf{0.5146} & \textbf{0.4328} & \textbf{0.4115} & \underline{0.0007} & \underline{0.3412} & \textbf{0.2511} \\
HEAL & 0.8376 & 0.1165 & \underline{0.5327} & 0.4956 & 0.7312 & 0.0194 & 0.3701 & 0.3736 & 0.3107 & 0.0001 & 0.1751 & 0.1620 \\
CodeFilling & 0.7590 & 0.1088 & 0.4941 & 0.4540 & 0.6458 & 0.0196 & 0.4284 & 0.3646 & 0.2142 & 0.0002 & 0.1561 & 0.1235 \\
GenComm & 0.8060 & 0.1104 & 0.5090 & 0.4751 & \underline{0.7471} & 0.0213 & \underline{0.4681} & \underline{0.4122} & \underline{0.3475} & 0.0002 & 0.2813 & 0.2097 \\
\midrule
HyDRA w CoAlign & 0.8206 & 0.1263 & 0.5016 & 0.4828 & 0.6657 & 0.0244 & 0.3631 & 0.3511 & 0.1824 & 0.0004 & 0.2181 & 0.1336 \\
\textbf{HyDRA w AG-PGO (Ours)} & \textbf{0.8427} & \underline{0.1662} & 0.5189 & \textbf{0.5093} & 0.7116 & \underline{0.0361} & 0.4235 & 0.3904 & 0.2917 & \underline{0.0007} & \textbf{0.3569} & \underline{0.2164} \\
\bottomrule
\end{tabular}%
}
\caption{Full baseline comparison under \textbf{pose noise} ($\sigma=0.4$) with \textbf{architecture heterogeneity} ($E_P + H_P + X_S + X_V$), extending Table~1(b) of the main paper.}
\label{tab:supp_robustness}
\end{table*}

\paragraph{Distribution analysis.}
The percentiles in \cref{tab:subsampling_spread} reveal the mechanism behind these error trends, which the per-frame means alone do not expose.
For homogeneous agents ($H_P$), the mean score stays above $\tau$ even at $K=1$, yet the $1$st percentile is $0.000$: under extreme sparsity the lower tail of the distribution collapses below $\tau$, which is what drives the high error rate. As $K$ grows, this lower tail lifts ($\text{p1}: 0.000 \rightarrow 0.183$) and crosses $\tau$ by $K=20$, matching the sharp drop in misclassification.
For latent-domain heterogeneous agents ($X_P$), the mean is essentially flat ($\approx 0.013$); here it is the upper tail that matters. The $99$th percentile shrinks monotonically ($0.550 \rightarrow 0.098$) and falls below $\tau$ at $K \geq 10$, explaining why these agents are reliably rejected once enough objects are shared.
Architecturally heterogeneous agents ($X_{S/V}$) are omitted, as their scores are identically $\approx 0$ ($\text{p1}=\text{p99}=0.000$) across all $K$.
Both regimes thus stabilize around $K=10$, consistent with the trend in the main paper.

\begin{table}[t]
\centering
\setlength{\tabcolsep}{6pt}
\resizebox{\columnwidth}{!}{%
\begin{tabular}{cc cccccc}
\toprule
\multirow{2}{*}{\textbf{Agent}} & \multirow{2}{*}{\textbf{$N$}} & \multicolumn{6}{c}{\textbf{Err. (\%) at $K$}} \\
\cmidrule(lr){3-8}
 & & \textbf{1} & \textbf{2} & \textbf{3} & \textbf{5} & \textbf{10} & \textbf{20} \\
\midrule
\multirow{5}{*}{$H_P$}
 & 1  & 41.94 & 29.58 & 22.74 & 13.14 & 2.31 & 0.03 \\
 & 2  & 30.43 & 18.22 & 12.14 & 5.01  & 0.35 & 0.00 \\
 & 3  & 22.85 & 11.43 & 6.72  & 2.20  & 0.07 & 0.00 \\
 & 5  & 12.29 & 5.08  & 2.42  & 0.53  & 0.00 & 0.00 \\
 & 10 & 3.02  & 0.85  & 0.25  & 0.02  & 0.00 & 0.00 \\
\midrule
\multirow{5}{*}{$X_P$}
 & 1  & 2.99 & 4.66 & 5.27 & 2.87 & 0.96 & 0.22 \\
 & 2  & 4.66 & 4.24 & 1.99 & 0.99 & 0.35 & 0.08 \\
 & 3  & 5.47 & 1.92 & 1.37 & 0.54 & 0.10 & 0.00 \\
 & 5  & 3.12 & 1.02 & 0.50 & 0.14 & 0.01 & 0.00 \\
 & 10 & 0.91 & 0.21 & 0.06 & 0.01 & 0.00 & 0.00 \\
\bottomrule
\end{tabular}%
}
\caption{Misclassification rate (\%) across accumulated frames $N \in \{1,2,3,5,10\}$ and subsampled shared objects $K$, at threshold $\tau=0.135$.}
\label{tab:supp_accumulation}
\end{table}

\paragraph{Multi-frame accumulation across $N$.}
\Cref{tab:supp_accumulation} 
extends the $N \in \{1, 2, 10\}$ accumulation results in Table~5 of the main paper to the finer grid $N \in \{1, 2, 3, 5, 10\}$.
The error rate decreases with $N$ at nearly every sparsity level: for $H_P$ at $K=1$, error falls from $41.94\%$ ($N{=}1$) to $22.85\%$ ($N{=}3$) to $3.02\%$ ($N{=}10$). The one exception is $X_P$ at $K=1$---the most extreme sparsity setting---where error rises transiently from $2.99\%$ ($N{=}1$) to $5.47\%$ ($N{=}3$) before dropping sharply to $0.91\%$ at $N{=}10$; we attribute this to sampling noise inherent to averaging over very few single-shared-object frames, an effect that increasing $N$ progressively averages out.

\paragraph{On the choice of $\tau$.}
The threshold $\tau=0.135$ is specific to this stress test and is not the deployment threshold. Under full observations the homogeneous and heterogeneous scores are cleanly separated, and any value within the feasible region (\cref{sec:supp_domain_score}, $[0.166,\,0.236]$) produces identical routing; the exact threshold is immaterial. Subsampling, however, shifts the homogeneous score distribution downward---its per-frame mean drops from $\approx\!0.51$ under full observations to $\approx\!0.24$--$0.35$---so the full-observation feasible region no longer applies and the two regimes partially overlap. We therefore adopt $\tau=0.135$ as the operating point for this sparse regime. Importantly, the conclusion we draw---that reliability degrades only under extreme sparsity and stabilizes by $K\approx 10$---reflects the shape of the distributions and is insensitive to
the precise threshold.

\section{Baseline Robustness under Pose Noise}
\label{sec:supp_robustness}

\Cref{tab:supp_robustness} reports the full pose-noise robustness comparison ($\sigma=0.4$) under architecture heterogeneity, extending Table~1(b) of the main paper. Due to space constraints, the table shows only HyDRA equipped with CoAlign and with our proposed AG-PGO; here we additionally report Late Fusion, E2E Training, and the four heterogeneity-aware baselines (MPDA, HEAL, CodeFilling, GenComm) under the identical noisy setting.

Most baselines degrade sharply at the strict AP@0.7 threshold relative to their noise-free performance in Table~1(a) of the main paper: Late Fusion drops from $0.2323$ to $0.0474$, HEAL from $0.3256$ to $0.1620$, and CodeFilling from $0.2626$ to $0.1235$, since these methods have no explicit mechanism to correct localization error before fusion. MPDA and GenComm degrade less severely and remain the strongest baselines under noise ($0.2511$ and $0.2097$ AP@0.7, respectively). HyDRA w/ AG-PGO remains competitive with these noise-robust baselines ($0.2164$ AP@0.7) while requiring no additional training, and clearly outperforms the standard CoAlign pose-graph baseline ($0.1336$), confirming that the anchor-guided formulation—rather than pose graph optimization in general—drives HyDRA's robustness.

\begin{figure}[t]
    \centering
    \includegraphics[width=1.0\linewidth]{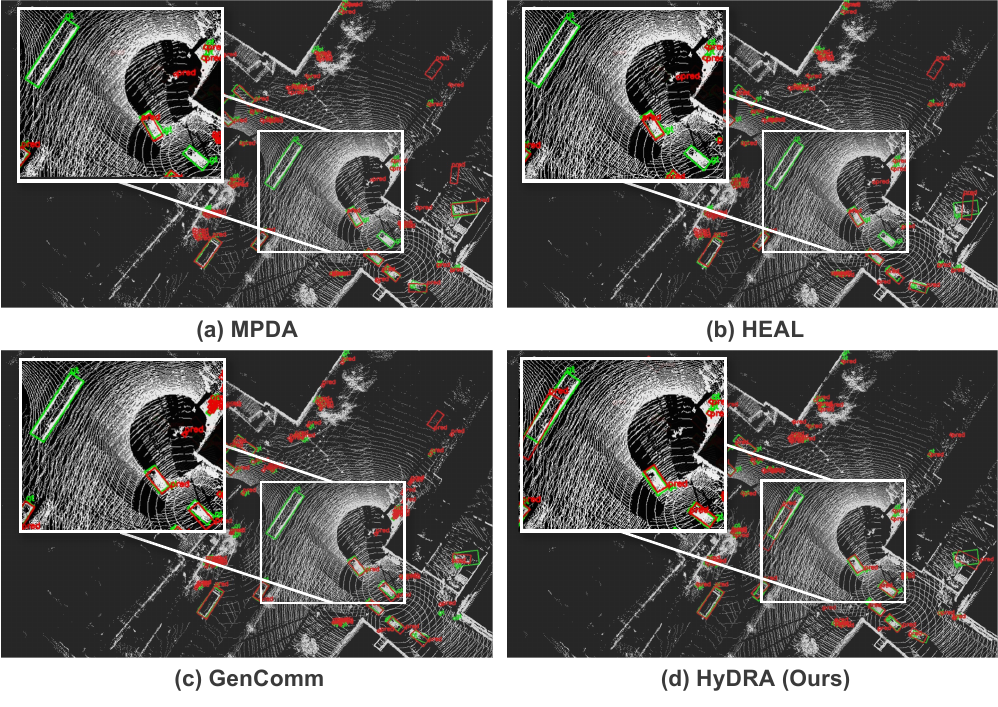}
    \caption{Qualitative comparison of 3D detection results with baseline methods. The \textcolor{red}{red} and \textcolor{green}{green} boxes represent the prediction and ground truth.}
    \label{fig:visualization}
\end{figure}

\section{Visualization}
\Cref{fig:visualization} presents the qualitative comparison of detection results between baseline methods and HyDRA. 
Baseline methods tend to suffer from false negatives, failing to identify several objects. 
In contrast, HyDRA recovers several objects missed by the baselines and produces fewer spurious detections.
While baselines also exhibit unstable predictions, our method maintains robust detection performance.

\section{Communication Overhead Analysis}
\label{sec:comm-overhead}
We quantify the per-agent, per-frame communication payload of HyDRA relative to the baselines, using raw intermediate-fusion feature transmission as the reference (100\%). The BEV feature map is $64$-channel in float32 for both datasets, but its spatial resolution differs by dataset due to differing detection ranges and voxel sizes: on OPV2V, the map is $256 \times 256$, totaling $64 \times 256 \times 256 \times 4~\text{B} \approx 16.78~\text{MB}$ per agent per frame; on V2X-Real, the map is $100 \times 352$, totaling $64 \times 100 \times 352 \times 4~\text{B} \approx 9.01~\text{MB}$ per agent per frame. All percentages below are computed relative to the reference of their respective dataset.

\paragraph{Baseline payloads.} HEAL and MPDA transmit the raw feature map without compression. GenComm transmits a $2$-channel compressed message and reconstructs the full feature at the receiver via diffusion, giving $2/64 = 3.125\%$ of the reference. CodeFilling transmits VQ codebook indices in place of feature vectors; its ablation study reports that this code index representation reduces communication cost by $208\times$ relative to transmitting the raw feature at matched detection performance, i.e.\ $\approx 0.48\%$ of the reference.

\begin{table}[t]
\centering
\small
\setlength{\tabcolsep}{8pt}
\begin{tabular}{llc}
\toprule
\textbf{Method} & \textbf{Payload} & \textbf{\% of Ref.} \\
\midrule
HEAL & feature & $100\%$ \\
MPDA & feature & $100\%$ \\
CodeFilling & VQ indices & $\approx 0.48\%$ \\
GenComm & 2-ch msg. & $3.125\%$ \\
\midrule
\multirow{4}{*}{HyDRA}
 & Onboard. (feature+det.) & $100\%{+}\epsilon^\dagger$ \\
 & Steady, hom. (feature) & $100\%$ \\
 & Steady, het. (det., V2X-Real) & $0.0132\%$ \\
 & Steady, het. (det., OPV2V) & $0.0032\%$ \\
\bottomrule
\end{tabular}
\caption{Per-agent, per-frame communication payload relative to raw intermediate-fusion feature transmission, computed per-dataset ($\approx 16.78$~MB on OPV2V, $\approx 9.01$~MB on V2X-Real). $^\dagger$$\epsilon \in [0.0032\%, 0.0132\%]$, negligible next to the feature cost.}
\label{tab:comm-overhead}
\end{table}

\paragraph{HyDRA payload.} HyDRA's transmitted payload depends on each agent's stage in the routing pipeline:
\begin{itemize}
\item \textbf{Onboarding}: the domain classifier requires both the feature map and the detection result every frame during this window, so both must be transmitted. The payload is therefore $100\% + \epsilon$, where $\epsilon \in [0.0032\%, 0.0132\%]$ is the detection-only overhead derived below; this window lasts up to $N=10$ frames under sparsity and as few as $N=1$ once $K \geq 10$ shared objects are observed.
\item \textbf{Steady-state, homogeneous ($\mathcal{N}_{int}$):} once routing is cached, only the feature map is needed for intermediate fusion; the detection result is no longer required. The payload remains $100\%$.
\item \textbf{Steady-state, heterogeneous ($\mathcal{N}_{late}$):}
once routing is cached, only the detection result is needed for late fusion; the feature map is no longer transmitted. Each detection comprises a 7-DoF box and a confidence score, totaling $32$B on OPV2V; V2X-Real additionally carries a class label across its three classes, totaling $33$B per detection. With the average number of detections per frame, $\overline{N}_{\text{gt}} = 36.12$ on V2X-Real and $\overline{N}_{\text{gt}} = 16.90$ on OPV2V, the per-frame payload is $36.12 \times 33~\text{B} \approx 1.19~\text{kB}$ on V2X-Real and $16.90 \times 32~\text{B} \approx 541~\text{B}$ on OPV2V.
\end{itemize}

\Cref{tab:comm-overhead} summarizes these figures. HyDRA's heterogeneous-agent steady-state payload is one to three orders of magnitude below GenComm and CodeFilling, at the cost of a transient full-payload onboarding window and a persistent full payload for homogeneous agents, which by construction benefit from feature-level fusion.

\end{document}

%% file: preamble.tex
\newcommand{\red}[1]{#1}

\usepackage{multirow}     %
\usepackage{colortbl}     %

\definecolor{idealBlue}{HTML}{E6F2FF}
\definecolor{robustRed}{HTML}{FFE6E6}

\usepackage{fontawesome5}
\usepackage[normalem]{ulem}
\usepackage[accsupp]{axessibility}

%% file: sec/0_abstract.tex
\begin{abstract}
Collaborative perception assumes that participating agents share compatible feature representations, yet this assumption often fails in practice. Beyond explicit differences in backbone architecture or sensor modality, independently trained models can exhibit latent domain shifts and cause naive feature fusion to contaminate rather than complement each agent's representation. To address this, we propose HyDRA (Hybrid Domain-Aware Robust Architecture), a training-free framework that uses a lightweight domain classifier to identify whether each agent shares a compatible feature domain, and adaptively select the fusion strategy. Homogeneous agents are combined through intermediate fusion to maximize information gain, whereas heterogeneous agents are routed to late fusion to avoid contamination. To further mitigate localization errors inherent in late fusion, HyDRA incorporates anchor-guided pose graph optimization. Extensive experiments show that HyDRA matches state-of-the-art methods under architecture heterogeneity, and improves overall AP@0.7 over late fusion by nearly 25\% under latent domain heterogeneity.
\end{abstract}

%% file: sec/1_intro.tex
\section{Introduction}
\label{sec:intro}

Collaborative perception (CP) enables Connected and Automated Vehicles (CAVs, hereafter referred to as ``agents'') to overcome the physical limitations of standalone sensing by sharing complementary information~\cite{chen2019cooper, xu2022opv2v, han2023collaborative, bae2024rethinking}. Despite its significant advantages, CP faces a critical challenge in real-world deployment: \textit{heterogeneity} among agents.

\begin{figure}[t]
    \centering
    \includegraphics[width=1.0\linewidth]{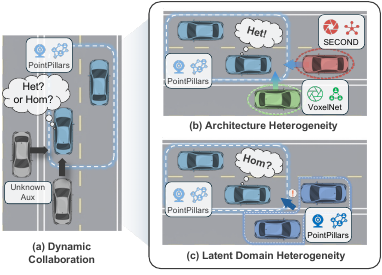}

\caption{Heterogeneity in dynamic collaboration. \red{\textbf{(a)} Previously unknown agents join the collaboration.} \textbf{(b) Architecture Heterogeneity}: diverse model architectures make direct feature fusion infeasible. \textbf{(c) Latent Domain Heterogeneity}: \red{independently trained models can exhibit} a subtler feature-domain shift even under identical architectures. \red{Our framework detects both from transmitted features.}}
    \label{fig:concept}
\end{figure}

As illustrated in Fig.~\ref{fig:concept}, neighboring agents inevitably differ in their sensing and learning configurations. Some mismatches are explicit and declarable: at setup, agents exchange metadata describing their configuration, including model architecture and sensor suite, so an architecture mismatch is evident from the start. 
\red{Others are latent: even under an identical architecture, independently trained models can produce diverging feature distributions, a divergence not captured by the architecture or dataset metadata exchanged at setup.} Prior heterogeneity-aware methods identify heterogeneity solely from declared configuration differences, casting the problem as aligning a mismatched agent to the ego domain. Latent domain heterogeneity escapes this view. An agent sharing the ego's architecture is treated as homogeneous and fused without alignment, 
\red{regardless of how far its feature distribution has drifted.}

\red{Intermediate fusion methods~\cite{xia2025one, gao2025stamp, luo2024plug, xu2023mpda, lu2024an, YueCodeFilling:CVPR2024, zhou2025pragmatic} maximize information gain by sharing features and can partially compensate for minor misalignment, but under latent domain heterogeneity, feature sharing may introduce contamination, and many such approaches require additional learning or adaptation when new agents join. Late fusion~\cite{fadili2025late, Chen2022ModelAgnosticMP} instead aggregates detections, a robust fallback under severe heterogeneity that avoids feature-space misalignment but sacrifices fine-grained information exchange and is more sensitive to localization noise.}

Therefore, we propose \textbf{HyDRA} (Hybrid Domain-Aware Robust Architecture), a training-free framework that identifies the domain compatibility of each collaborating agent at inference time and adaptively selects its fusion strategy. 
\red{At its core is a lightweight domain classifier that probes whether the ego's frozen decoder can interpret each agent's transmitted features, requiring no metadata about the agent's architecture or training beyond a common BEV feature interface.} The resulting per-agent decision drives the fusion strategy: agents judged homogeneous are integrated via intermediate fusion to maximize feature-level information sharing, while those judged heterogeneous are fused at the detection level via late fusion to avoid feature contamination. To mitigate the localization-noise sensitivity inherent to late fusion, we further introduce an anchor-guided pose graph optimization (AG-PGO) that refines the poses of heterogeneous agents.
\red{Since onboarding a new agent requires no training or parameter updates, only a 9.4\,ms evaluation per agent per frame over up to $10$ frames, HyDRA is well suited to open environments where agents dynamically join and leave.}

\begin{figure*}[t]
    \centering
    \includegraphics[width=1.0\linewidth]{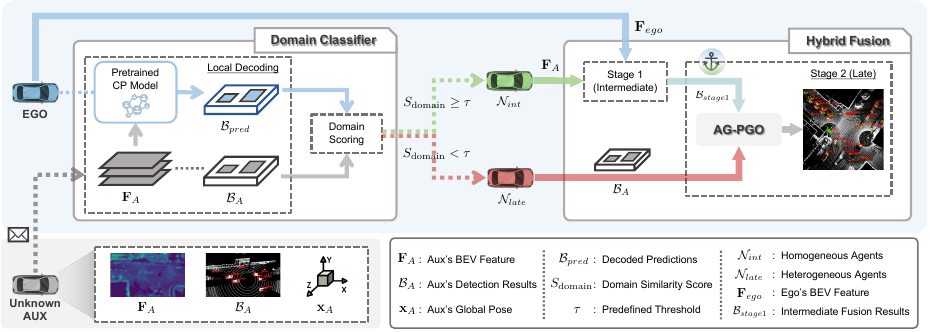}
    \caption{Overview of \textbf{HyDRA}. The \textbf{Domain Classifier} scores each agent's compatibility via $S_{\text{domain}}$; homogeneous agents ($\mathcal{N}_{int}$) undergo \textbf{Intermediate Fusion} to produce Stage 1 detections ($\mathcal{B}_{stage1}$), while heterogeneous agents ($\mathcal{N}_{late}$) undergo \textbf{Late Fusion}. \textbf{AG-PGO} then corrects heterogeneous agents' poses using $\mathcal{B}_{stage1}$ as fixed spatial anchors.}
    \label{fig:framework}
\end{figure*}
 
Our contributions are summarized as follows:

\red{\begin{itemize}
    \item We characterize latent domain heterogeneity, where agents sharing the ego's architecture and training dataset but trained independently can be feature-incompatible in ways metadata does not reveal, and show that naively fusing them degrades performance.
    \item We propose a lightweight domain classifier that probes feature compatibility at inference time using the ego's frozen decoder, without metadata about the agent's architecture or training and without retraining.
    \item We propose HyDRA, a hybrid fusion framework that routes each agent to intermediate or late fusion based on the classifier to avoid feature contamination while maximizing information gain, and exploits this two-stage structure through AG-PGO, which adapts agent--object pose graph optimization
to use intermediate-fusion detections as fixed anchors for the late-fusion branch.
    \item HyDRA matches state-of-the-art heterogeneity-aware methods under architecture heterogeneity without additional training, and improves AP@0.7 over late fusion by nearly 25\% under latent domain heterogeneity.
\end{itemize}}

%% file: sec/2_problem.tex
\section{Problem Statement}
\label{Sec:Problem}

When previously unknown agents dynamically join a CP (\cref{fig:concept}(a)), heterogeneity becomes inevitable. Due to differences in model design and training environment, incoming information from neighboring agents may not be directly compatible with the ego agent. Such mismatches introduce domain shifts that can disrupt effective collaboration.

In dynamic CP, heterogeneity manifests in two forms. The first is \textit{architecture heterogeneity} (\cref{fig:concept}(b)), where agents employ different backbone networks, making direct feature fusion structurally infeasible. In many prior works, such architectural differences are declarable through metadata exchanged at setup, so the corresponding agents can be routed accordingly.
The second is \textit{latent domain heterogeneity} (\cref{fig:concept}(c)), which is more subtle. 

\red{Even when agents share the same model architecture, independently trained models may produce feature representations with domain shifts. Here, independently trained means trained separately from the ego rather than jointly with it; in our experiments, this also entails a different training regime (single-agent vs.\ joint collaborative training; \cref{subsec:setup}). Unlike FDA~\cite{li2024breaking}, which targets domain gaps arising from different training datasets, we consider shifts that arise under an identical dataset and are therefore not revealed by architecture or dataset metadata.}
Thus, neither architectural nor dataset compatibility guarantees feature-level consistency, and blindly fusing such features can degrade perception performance.

This paper aims to design a robust CP framework for dynamic environments with heterogeneous agents. Specifically, the objective is to maximize information gain from homogeneous agents while preventing performance degradation caused by heterogeneous ones.

%% file: sec/3_hydra.tex
\section{Hybrid Domain-Aware Robust Architecture}
\label{Sec:Method}

\red{HyDRA consists of three components (\cref{fig:framework}): a domain classifier that assesses each agent's feature compatibility using the ego's frozen CP model as a probe, a hybrid fusion module that fuses compatible agents at the feature level and the rest at the detection level, and AG-PGO, which refines heterogeneous agents' poses using intermediate-fusion detections as fixed anchors. In the following subsections, we detail each component.}

\subsection{Domain Classifier}
\label{subsec:domain_classifier}

The guiding principle of our domain classifier is to use the ego's own frozen perception model as a probe for domain compatibility. Intuitively, if the ego can decode an agent's transmitted features into detections that agree with the agent's own outputs, the two operate in a compatible feature domain; 
\red{otherwise, the features are incompatible with the ego's decoder.} This turns compatibility assessment into an inference-time measurable quantity that requires no labels, no shared metadata, and no additional training. 
Note that this consistency signal targets benign but incompatible agents rather than malicious ones; unlike CP-Guard~\cite{hu2025cp}, which uses prediction-consistency to detect adversarial agents, our classifier assumes cooperative agents and instead surfaces unintentional domain shift.

Consider an auxiliary agent $A$ that transmits a tuple $(\mathbf{F}_A, \mathcal{B}_A, \mathbf{x}_A)$ to the ego agent.
Here $\mathbf{F}_A$ denotes the intermediate spatial feature map, $\mathcal{B}_A = \{(\mathbf{b}^i_A, c^i_A)\}_{i=1}^{N_{\text{gt}}}$ represents the agent's detection results, with $N_{\text{gt}}$ denoting the total number of detected objects, and $\mathbf{x}_A$ denotes the global pose of the agent. Each detection comprises a 7-DoF 3D bounding box $\mathbf{b}^i_A \in \mathbb{R}^7$ (position, size, orientation) and a confidence score $c^i_A \in [0, 1]$. 

The domain classifier evaluates the domain similarity between each auxiliary agent and the ego agent at inference time in three steps: (1) Local decoding, (2) Pairwise quality estimation, and (3) Soft-AP domain scoring.

\textbf{Step 1: Local Decoding.} The ego agent processes the received features through its own frozen pre-trained CP model to obtain locally decoded predictions: $\mathcal{B}_{\text{pred}} = \text{Decode}(\mathbf{F}_A) = \{(\mathbf{b}^k_{\text{pred}}, c^k_{\text{pred}})\}_{k=1}^{N_{\text{pred}}}$. $\text{Decode}(\cdot)$ denotes the ego's frozen fusion and detection-head modules.
Here, $N_{\text{pred}}$ is the total number of local predictions, and $\mathbf{b}^k_{\text{pred}}$ and $c^k_{\text{pred}}$ denote the $k$-th predicted 3D bounding box and its confidence score, respectively.

\red{Following standard heterogeneous CP pipelines~\cite{lu2024an}, we assume a common feature interface: agents share a detection range, BEV cell size, and channel dimension, so that $\mathbf{F}_A$ is dimensionally and geometrically compatible with the ego's decoder. Each agent meets this interface by setting its encoder's voxel size and backbone stride accordingly. This is a protocol-level agreement, not per-agent metadata on the sender's architecture or training.}

These predictions reveal how well the ego's decoder can interpret the incoming features. 
Since ground truth annotations are unavailable, we treat $\mathcal{B}_A$ as \textit{pseudo-ground truths}, 
\red{a reference for what the sender's features encode rather than a measure of their correctness.} We thus establish one-to-one correspondences between $\mathcal{B}_{\text{pred}}$ and $\mathcal{B}_A$ using the Hungarian algorithm with IoU as the matching cost.
This matching process yields a set of matched index pairs $\mathcal{M} = \{(i, k)\}$, linking the $i$-th pseudo-ground truth in $\mathcal{B}_A$ to the $k$-th prediction in $\mathcal{B}_{\text{pred}}$.

\textbf{Step 2: Pairwise Quality Estimation.} For each matched pair $(i, k) \in \mathcal{M}$, we compute a quality score $q_k \in [0, 1]$ that jointly captures spatial accuracy and semantic consistency. Specifically, $q_k$ is the geometric mean 
$q_k = \sqrt{S_{\text{conf}} \cdot S_{\text{iou}}}$, where the semantic consistency is $S_{\text{conf}} = \exp\left(-\frac{|c^i_A - c^k_{\text{pred}}|}{\sigma}\right)$ and the spatial accuracy is $S_{\text{iou}} = \text{IoU}(\mathbf{b}^i_A, \mathbf{b}^k_{\text{pred}})$. Here, $\sigma$ is a temperature parameter controlling the sensitivity to confidence discrepancies. 
This design ensures that achieving a high quality score requires both precise localization (high IoU) and preserved semantic confidence. 
Unmatched predictions—representing either false positives from the ego decoder or missed detections—are assigned $q_k = 0$.

\textbf{Step 3: Soft-AP Domain Scoring.} Traditional Average Precision (AP) determines true positives using binary matching criteria (e.g., fixed IoU thresholds), which are insufficient for capturing the gradual and continuous nature of domain shifts. To address this limitation, we define a Soft Average Precision (Soft-AP) formulation that replaces binary hit/miss decisions with continuous quality scores.

To compute this, we first sort the predictions in descending order of their confidence scores $c^k_{\text{pred}}$. Based on this sorted order, we progressively accumulate the corresponding quality scores $q_m$ and compute soft precision and recall at each rank $m$:
$P_m = \frac{\sum_{j=1}^{m} q_j}{m}, \quad R_m = \frac{\sum_{j=1}^{m} q_j}{N_{\text{gt}}} $. Here, $P_m$ represents the average quality of the top-$m$ predictions (soft precision), while $R_m$ measures the cumulative quality relative to the total number of pseudo-ground truths (soft recall). 
The domain similarity score $S_{\text{domain}}$ is then computed as the area under this soft precision-recall curve.

\red{We validate this design against mean-IoU and Hard-AP alternatives (\cref{subsec:dc_analysis}; supp.\ Secs.~B,~E).}
Based on these scores, we dynamically partition the auxiliary agents $\mathcal{N}$ into two disjoint groups: $ \mathcal{N}_{int} = \{j \mid S^j_{\text{domain}} \geq \tau\}, \quad \mathcal{N}_{late} = \{j \mid S^j_{\text{domain}} < \tau\}$, where $\tau$ is a predefined threshold. Agents in $\mathcal{N}_{int}$ are classified as homogeneous, and agents in $\mathcal{N}_{late}$ are treated as heterogeneous. 
This classification is cached for subsequent frames once committed. With at least $K = 10$ shared objects in the first frame, the single-frame decision is already reliable---
\red{misclassification stays below $2.5\%$ for every agent type (\cref{tab:subsampling})---}and is committed immediately. Below this, the commitment is deferred until scores have been accumulated over $N = 10$ frames, which drives the misclassification rate down to a few percent or below in nearly all cases.
While a decision is pending, the agent is provisionally routed to late fusion.

\subsection{Hybrid Fusion}
\label{subsec:hybrid_fusion}

Unlike conventional approaches that enforce a unified fusion for all agents, we structure the process sequentially: (1) intermediate fusion for homogeneous agents, followed by (2) late fusion that integrates the intermediate results with the independent predictions of heterogeneous agents. 

\textbf{Stage 1: Selective Intermediate Fusion.}
We aim to fully exploit the spatial information from agents in $\mathcal{N}_{int}$.
Since their feature representations are compatible with the ego agent, direct feature-level fusion can enhance perception performance without the risk of feature contamination.
Let $\Phi_{int}(\cdot)$ denote the intermediate fusion operator, such as Pyramid Fusion~\cite{lu2024an}. We fuse the ego feature map $\mathbf{F}_{ego}$ with the feature maps $\{ \mathbf{F}_j \mid j \in \mathcal{N}_{int} \}$ as $\mathcal{B}_{stage1} = \Phi_{int} \left( \{\mathbf{F}_{ego}\} \cup \{ \mathbf{F}_j \mid j \in \mathcal{N}_{int} \} \right)$.
The resulting detection set $\mathcal{B}_{stage1}$ provides spatially reliable and semantically consistent bounding boxes, which serve as fixed spatial anchors for the subsequent pose graph optimization.

\textbf{Stage 2: Late Fusion.} 
For the heterogeneous group $\mathcal{N}_{late}$, we adopt a late-fusion strategy that operates directly on detection outputs, thereby preserving the independence of each agent’s prediction and avoiding feature-level interference. 
Let $\Phi_{late}(\cdot)$ denote the late-fusion operator, which aggregates detection sets from multiple agents. We aggregate the first-stage predictions $\mathcal{B}_{stage1}$ with the detection sets $\{\mathcal{B}_j \mid j \in \mathcal{N}_{late}\}$ as $\mathcal{B}_{final} = \Phi_{late} \left( \mathcal{B}_{stage1} \cup \bigcup_{j \in \mathcal{N}_{late}} \mathcal{B}_j \right).$

\subsection{Anchor-Guided Pose Graph Optimization}
\label{subsec:pgo}
\begin{figure}[t]
    \centering
    \includegraphics[width=1.0\linewidth]{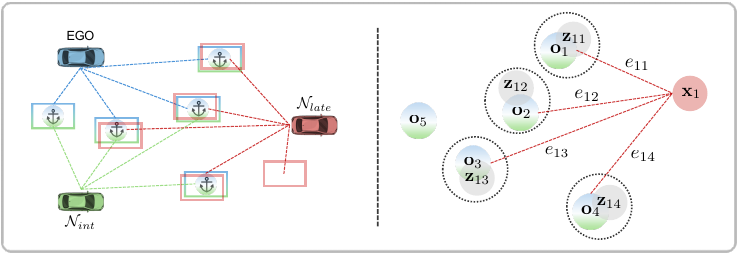}
    \caption{\textbf{AG-PGO}.
    Instead of global optimization, we leverage the reliable detections from homogeneous agents ($\mathcal{N}_{int}$) as fixed spatial anchors (\faAnchor).
    The module corrects the pose of heterogeneous agents ($\mathcal{N}_{late}$) by minimizing the residual errors between their predictions and the anchors, 
\red{improving consistency with the anchors.}}
    \label{fig:AG_PGO}
\end{figure}

Unlike intermediate fusion, where high-dimensional feature interactions can implicitly compensate for minor misalignment, late fusion depends heavily on the accurate global poses of participating agents to project bounding boxes into a common coordinate system. As a result, localization noise in heterogeneous agents can directly degrade fusion quality. To mitigate spatial misalignment in the late fusion, we propose an anchor-guided pose graph optimization (AG-PGO) module, conceptually illustrated in \cref{fig:AG_PGO}.

\subsubsection{Graph Construction}
We construct a graph $\mathcal{G} = (\mathcal{V}, \mathcal{E})$ where each node encodes a pose consisting of the 2D position and yaw angle. The node set $\mathcal{V}$ is composed of two types of nodes: variable pose nodes $\mathcal{X}$ for misaligned agents and fixed object anchors $\mathcal{O}$ corresponding to reliable detections.

\begin{itemize}
    \item \textbf{Variable Pose Nodes ($\mathcal{X}$):} 
    Each heterogeneous agent $i \in \mathcal{N}_{late}$ is represented by a variable pose node $\mathbf{x}_i = (x_i, y_i, \theta_i)$.
    The set $\mathcal{X} = \{\mathbf{x}_i \mid i \in \mathcal{N}_{late}\}$ defines the decision variables to be optimized. 
\red{The ego agent and homogeneous agents are excluded from $\mathcal{X}$: anchors are expressed in the ego frame, and pose errors of homogeneous agents are not corrected explicitly but are partially compensated by feature-level fusion in Stage~1. Without any homogeneous agents, the ego's own detections serve as anchors.}
    
    \item \textbf{Fixed Object Anchors ($\mathcal{O}$):} 
    We use the Stage~1 detections $\mathcal{B}_{stage1}$ to represent fixed object anchors.
    For the $k$-th detection in $\mathcal{B}_{stage1}$, we extract its 2D center coordinates $(x_k, y_k)$ and yaw angle $\theta_k$ from the 3D bounding box to define the anchor pose $\mathbf{o}_k = (x_k, y_k, \theta_k)\in \mathcal{O}$.
    These anchors remain fixed during optimization and serve as rigid spatial constraints in the pose graph.
\end{itemize}

We establish an edge $e_{ik} \in \mathcal{E}$ between the variable pose node $\mathbf{x}_i$ and the fixed anchor $\mathbf{o}_k$ when a detection from heterogeneous agent $i$ is matched to anchor $k$ based on predefined distance and yaw difference thresholds. 
Upon a successful match, we define the observation $\mathbf{z}_{ik} = (x_{ik}, y_{ik}, \theta_{ik})$, which denotes agent $i$’s estimated pose of object $k$ in its (uncorrected) local coordinate frame. This edge induces a geometric constraint: the global anchor $\mathbf{o}_k$, when transformed into agent $i$'s local coordinate system via the optimization variable $\mathbf{x}_i$, should align with the local observation $\mathbf{z}_{ik}$.

\subsubsection{Optimization with Confidence-aware Constraints}
Given the constructed graph, our goal is to find the optimal poses $\mathcal{X}^*$ of the heterogeneous agents that minimize the spatial discrepancy between their local observation $\mathbf{z}_{ik}$ and the projected global anchors $\mathbf{o}_k$.
We formulate the alignment as a nonlinear least-squares problem. 
\begin{equation}
    \mathcal{X}^* = \mathop{\mathrm{argmin}}_{\mathcal{X}} \sum_{(i, k):e_{ik} \in \mathcal{E}} \left\| \mathbf{r}_{ik}(\mathbf{x}_i, \mathbf{o}_k) \right\|_{\mathbf{\Omega}_{ik}}^2, \label{eq:optimization}
\end{equation}
where $\mathbf{r}_{ik} := \mathbf{z}_{ik} \ominus h(\mathbf{x}_i, \mathbf{o}_k)$ and $\mathbf{\Omega}_{ik} := (c_{aux, i})^{\gamma} \cdot (c_{anchor, k})^{\beta} \cdot \mathbf{I}$. Here, $\mathbf{r}_{ik}$ is the residual vector between the measurement $\mathbf{z}_{ik}$ and the anchor $\mathbf{o}_k$ transformed into agent $i$'s frame via $h(\cdot)$. The information matrix $\mathbf{\Omega}_{ik}$ is defined in terms of $c_{aux, i}$ and $c_{anchor, k}$, which are the confidence scores of the agent $i$'s detection and the anchor detection, respectively. Note that to enhance robustness against localization noise and false positives, we adopt a reliability-aware weighting scheme based on these confidence scores. The hyperparameters $\gamma$ and $\beta$ control the influence of each term, allowing low-confidence matches to be down-weighted while emphasizing reliable correspondences for stable pose correction.

We draw inspiration from the agent--object pose graph formulation introduced in CoAlign~\cite{lu2023robust}, which aligns coordinate frames using relative observations between agents and detected objects.
\red{AG-PGO is a targeted modification of this formulation for our hybrid pipeline.}
As depicted in \cref{fig:AG_PGO}, instead of jointly optimizing all auxiliary agents and detected objects, which may propagate localization errors across the entire pose graph and degrade the overall alignment, we restrict optimization to heterogeneous agents only. Specifically, the reliable detections $\mathcal{B}_{stage1}$, obtained through intermediate fusion of homogeneous agents, are treated as fixed spatial anchors, and only the poses of heterogeneous agents are modeled as variable nodes to be optimized. 
Importantly, our formulation relies solely on detections and poses from heterogeneous agents, without requiring specialized auxiliary networks.

%% file: sec/4_experiments.tex
\section{Experiments}\label{Sec:Experiments}
In this section, we evaluate HyDRA against baselines under heterogeneity, analyze scalability and efficiency, and present component-level and ablation analyses.

\subsection{Setup}

\red{We evaluate on V2X-Real~\cite{xiang2024v2x}, a large-scale real-world dataset with vehicles and roadside units (RSUs) and three classes (vehicle, pedestrian, truck), and OPV2V~\cite{xu2022opv2v}, a synthetic CARLA-based benchmark with connected vehicles and a single class (vehicle).}

\begin{table*}[t]
\centering
\resizebox{\textwidth}{!}{%
\begin{tabular}{l cccc cccc cccc}
\toprule
\multirow{2}{*}{\textbf{Method}} & \multicolumn{4}{c}{\textbf{AP@0.3}} & \multicolumn{4}{c}{\textbf{AP@0.5}} & \multicolumn{4}{c}{\textbf{AP@0.7}} \\
\cmidrule(lr){2-5}\cmidrule(lr){6-9}\cmidrule(lr){10-13}
 & \textbf{vehicle} & \textbf{pedestrian} & \textbf{truck} & \textbf{total} & \textbf{vehicle} & \textbf{pedestrian} & \textbf{truck} & \textbf{total} & \textbf{vehicle} & \textbf{pedestrian} & \textbf{truck} & \textbf{total} \\
\midrule
\rowcolor{idealBlue}
\multicolumn{13}{c}{\textbf{(a) Ideal Setting (No Noise)}} \\
\midrule
No Fusion & 0.6471 & 0.3285 & 0.5161 & 0.4972 & 0.6006 & 0.1248 & 0.4300 & 0.3851 & 0.3208 & 0.0016 & 0.2306 & 0.1843 \\
Late Fusion & 0.7297 & \underline{0.4327} & 0.4898 & 0.5507 & 0.7096 & \underline{0.1742} & 0.4167 & 0.4335 & 0.4241 & \textbf{0.0073} & 0.2654 & 0.2323 \\
E2E Training & 0.8641 & 0.4087 & 0.5542 & 0.6090 & \underline{0.8438} & 0.1444 & 0.5038 & 0.4974 & 0.5693 & 0.0031 & 0.2547 & 0.2757 \\
MPDA~\cite{xu2023mpda} & 0.8288 & 0.3790 & \underline{0.5645} & 0.5908 & 0.8061 & 0.1463 & \underline{0.5426} & 0.4983 & 0.5241 & 0.0045 & 0.3850 & 0.3045 \\
HEAL~\cite{lu2024an} & 0.8661 & 0.4115 & \textbf{0.6100} & \textbf{0.6292} & 0.8392 & 0.1553 & 0.5348 & \textbf{0.5098} & \underline{0.5706} & 0.0040 & 0.4023 & 0.3256 \\
CodeFilling~\cite{YueCodeFilling:CVPR2024} & 0.8078 & 0.3483 & 0.5364 & 0.5641 & 0.7638 & 0.1219 & 0.4959 & 0.4605 & 0.4373 & 0.0031 & 0.3473 & 0.2626 \\
GenComm~\cite{zhou2025pragmatic} & \textbf{0.8963} & 0.3602 & 0.5599 & 0.6054 & 0.8396 & 0.1282 & \textbf{0.5440} & 0.5039 & 0.5349 & 0.0040 & \textbf{0.4528} & \underline{0.3306} \\
\midrule
\textbf{HyDRA (Ours)} & \underline{0.8684} & \textbf{0.4416} & 0.5423 & \underline{0.6174} & \textbf{0.8555} & \textbf{0.1818} & 0.4884 & \underline{0.5086} & \textbf{0.5914} & \underline{0.0059} & \underline{0.4078} & \textbf{0.3350} \\
\midrule
\rowcolor{robustRed}
\multicolumn{13}{c}{\textbf{(b) Robustness Analysis (Gaussian Pose Noise: $\sigma=0.4$)}} \\
\midrule
HyDRA w CoAlign~\cite{lu2023robust} & 0.8206 & 0.1263 & 0.5016 & 0.4828 & 0.6657 & 0.0244 & 0.3631 & 0.3511 & 0.1824 & 0.0004 & 0.2181 & 0.1336 \\
\textbf{HyDRA w AG-PGO (Ours)} & 0.8427 & 0.1662 & 0.5189 & 0.5093 & 0.7116 & 0.0361 & 0.4235 & 0.3904 & 0.2917 & 0.0007 & 0.3569 & 0.2164 \\
\bottomrule
\end{tabular}%
}
\caption{Performance comparison under \textbf{architecture heterogeneity}: $E_P + H_P + X_S + X_V$}
\label{tab:results}
\end{table*}

\begin{figure*}[t]
    \centering
    \includegraphics[width=1.0\linewidth]{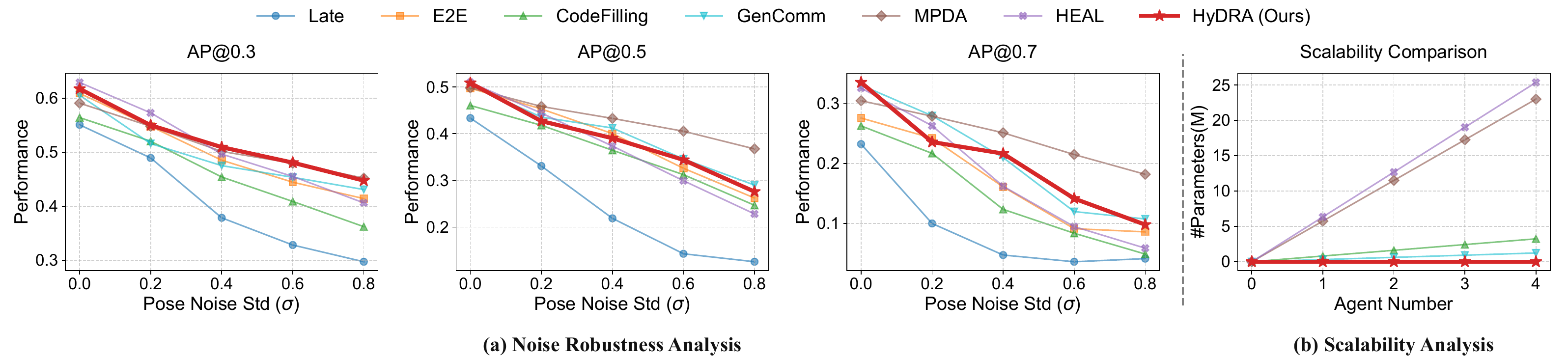}
    \caption{Performance comparison with baseline methods under varying pose noise and scalability analysis}
    \label{fig:noise}
\end{figure*}

\begin{table}[t]
\centering
\small
\setlength{\tabcolsep}{4pt}
\resizebox{\columnwidth}{!}{%
\begin{tabular}{l cc cc cc}
\toprule
\multirow{2}{*}{\textbf{Method}} & \multicolumn{2}{c}{\textbf{AP@0.3}} & \multicolumn{2}{c}{\textbf{AP@0.5}} & \multicolumn{2}{c}{\textbf{AP@0.7}} \\
\cmidrule(lr){2-3}\cmidrule(lr){4-5}\cmidrule(lr){6-7}
 & \textbf{A} & \textbf{L} & \textbf{A} & \textbf{L} & \textbf{A} & \textbf{L} \\
\midrule
MPDA~\cite{xu2023mpda} & 0.8796 & 0.8763 & 0.8731 & 0.8696 & 0.8191 & 0.8075 \\
HEAL~\cite{lu2024an} & \underline{0.9102} & \underline{0.9017} & \underline{0.8978} & \underline{0.8910} & \textbf{0.8395} & \textbf{0.8336} \\
CodeFilling~\cite{YueCodeFilling:CVPR2024} & 0.8865 & 0.8864 & 0.8747 & 0.8725 & 0.7950 & 0.7903 \\
GenComm~\cite{zhou2025pragmatic} & 0.8797 & 0.8575 & 0.8736 & 0.8494 & 0.7985 & 0.7766 \\
\midrule
\textbf{HyDRA (Ours)} & \textbf{0.9156} & \textbf{0.9181} & \textbf{0.9014} & \textbf{0.9023} & \underline{0.8238} & \underline{0.8213} \\
\bottomrule
\end{tabular}%
}
\caption{Performance comparison on \textbf{OPV2V} under architecture heterogeneity (\textbf{A}) and latent domain heterogeneity (\textbf{L}).}
\label{tab:opv2v_combined}
\end{table}

\subsubsection{Heterogeneous Settings}
\label{subsec:setup}
To empirically validate our framework under the dynamic collaboration illustrated in \cref{fig:concept}(a), we construct a realistic collaborative environment involving four distinct agents.
We denote the type of each agent using the notation $T_b$, where $T \in \{E, H, X\}$ represents the agent type (Ego, Homogeneous, Heterogeneous) and $b \in \{P, S, V\}$ denotes the backbone architecture (PointPillars~\cite{lang2019pointpillars}, SECOND~\cite{yan2018second}, VoxelNet~\cite{zhou2018voxelnet}).
The specific roles and configurations are defined as follows:

\begin{itemize}
    \item \textbf{Ego Agent ($E_P$):} The ego agent utilizes \textit{PointPillars} ($P$) as its 3D object detection encoder.
    
    \item \textbf{Homogeneous Auxiliary Agent ($H_P$):} This agent acts as a baseline for ideal collaboration. It shares the exact \textit{PointPillars} architecture and 
\red{is jointly trained with the ego agent ($E_P$), so that the two share a feature domain by construction.}
    
    \item \textbf{Heterogeneous Auxiliary Agents ($X$):} To evaluate robustness against the two heterogeneity types defined in \cref{Sec:Problem}, we introduce:
    \begin{itemize}
        \item \textit{Architecture Heterogeneity ($X_S, X_V$):} Corresponding to \cref{fig:concept}(b), these agents utilize structurally distinct backbones (\textit{SECOND} or \textit{VoxelNet}). This setting simulates the explicit model mismatch scenario.
        
        \item \textit{Latent Domain Heterogeneity ($X_P$):} To reproduce the threat depicted in \cref{fig:concept}(c), we include an agent that uses the same PointPillars architecture as the ego and is trained on the same dataset, 
\red{but independently, in a single-agent setting, rather than jointly with $E_P$. $X_P$ thus matches $H_P$ in architecture and dataset metadata but differs in both the training run and the training regime; we do not disentangle the two factors.}
    \end{itemize}
\end{itemize}

\subsubsection{Implementation Details}
We assume that within each agent, the same backbone network is used to generate both its feature map and single-agent perception outputs.
Distinct voxel sizes are configured for each encoder.
Specifically, on V2X-Real the voxel size is set to $[0.4, 0.4, 30]$ for PointPillars. 
We set the voxel size to $[0.1, 0.1, 0.1]$ for SECOND and $[0.4, 0.4, 3]$ for VoxelNet, all in meters.
To integrate the features from these encoders, we employ Pyramid Fusion~\cite{lu2024an} as our feature fusion method.
The detection range is defined as $[-140.8, 140.8]\,\text{m}$ along the $x$-axis, $[-40, 40]\,\text{m}$ along the $y$-axis, and $[-15, 15]\,\text{m}$ along the $z$-axis.
On OPV2V, only the vertical voxel extent and the detection range differ: $4$ and $0.8$ for the $z$-axis of PointPillars and VoxelNet, over a range of $[-102.4, 102.4]$ along the $x$- and $y$-axes and $[-3, 1]$ along the $z$-axis, again in meters.
For fair comparison, all baselines are reproduced. 

We train each baseline for 25 epochs using the AdamW~\cite{Loshchilov2017DecoupledWD} optimizer with a unified batch size of 2 on a single NVIDIA RTX 4090 GPU. \red{For pose noise, we perturb the $x$, $y$ position and yaw of every agent with zero-mean Gaussian noise of standard deviation $\sigma$ (in m and $^\circ$, respectively).}

\begin{table*}[t]
\centering
\resizebox{\textwidth}{!}{%
\begin{tabular}{l cccc cccc cccc}
\toprule
\multirow{2}{*}{\textbf{Method}} & \multicolumn{4}{c}{\textbf{AP@0.3}} & \multicolumn{4}{c}{\textbf{AP@0.5}} & \multicolumn{4}{c}{\textbf{AP@0.7}} \\
\cmidrule(lr){2-5}\cmidrule(lr){6-9}\cmidrule(lr){10-13}
 & \textbf{vehicle} & \textbf{pedestrian} & \textbf{truck} & \textbf{total} & \textbf{vehicle} & \textbf{pedestrian} & \textbf{truck} & \textbf{total} & \textbf{vehicle} & \textbf{pedestrian} & \textbf{truck} & \textbf{total} \\
\midrule
Intermediate Fusion & 0.6492 & 0.2531 & 0.3628 & 0.4217 & 0.6236 & 0.1119 & 0.3553 & 0.3636 & 0.4166 & 0.0087 & 0.2897 & 0.2383 \\
Late Fusion & 0.7078 & 0.3348 & 0.4663 & 0.5030 & 0.6781 & 0.1259 & 0.4350 & 0.4130 & 0.4350 & 0.0083 & 0.2917 & 0.2450 \\
MPDA~\cite{xu2023mpda} & \underline{0.7985} & \textbf{0.3778} & 0.5313 & 0.5692 & \underline{0.7801} & \underline{0.1572} & 0.4945 & \underline{0.4773} & \underline{0.5691} & \textbf{0.0093} & 0.3128 & 0.2971 \\
HEAL~\cite{lu2024an} & 0.7983 & 0.2909 & 0.4914 & 0.5269 & 0.7757 & 0.1121 & 0.4349 & 0.4409 & 0.5408 & 0.0054 & 0.2780 & 0.2747 \\
CodeFilling~\cite{YueCodeFilling:CVPR2024} & 0.7681 & \underline{0.3748} & \textbf{0.5860} & \underline{0.5763} & 0.7524 & \textbf{0.1577} & \textbf{0.5358} & \textbf{0.4820} & \textbf{0.5724} & 0.0082 & \textbf{0.4101} & \textbf{0.3302} \\
GenComm~\cite{zhou2025pragmatic} & 0.7662 & 0.3107 & 0.4933 & 0.5234 & 0.7443 & 0.1250 & 0.4632 & 0.4442 & 0.5364 & 0.0067 & 0.3271 & 0.2901 \\
\midrule
\textbf{HyDRA (Ours)} & \textbf{0.8171} & 0.3712 & \underline{0.5619} & \textbf{0.5834} & \textbf{0.7882} & 0.1452 & \underline{0.4961} & 0.4765 & 0.5612 & \underline{0.0092} & \underline{0.3479} & \underline{0.3061} \\
\bottomrule
\end{tabular}%
}
\caption{Performance comparison under \textbf{latent domain heterogeneity} ($E_P + H_P + X_P + X_P$). All heterogeneity-aware baselines are equipped with our domain classifier for routing.}
\label{tab:res}
\end{table*}

\subsection{Analysis of Architecture Heterogeneity}
\label{subsec:architectural_analysis}

We first evaluate the performance under explicit architectural mismatch.
\Cref{tab:results}(a) presents the quantitative comparison of our proposed HyDRA against baseline approaches. 
The experiments are conducted under the architecture heterogeneity setting ($E_P + H_P + X_S + X_V$).
To comprehensively evaluate our framework, we compare it against heterogeneity-aware methods, including MPDA, HEAL, CodeFilling, and GenComm~\cite{xu2023mpda, lu2024an, YueCodeFilling:CVPR2024, zhou2025pragmatic}. E2E (end-to-end) training jointly optimizes the entire collaborative perception pipeline across different domains.

\red{Our method achieves the highest total AP@0.7, narrowly ahead of GenComm (0.3350 vs.\ 0.3306), while} HEAL scores marginally higher at lower IoU thresholds; per-class results follow a similar trend.
We observe a consistent trend on OPV2V (\cref{tab:opv2v_combined}, \textbf{A} columns): HyDRA achieves the best performance at AP@0.3 and AP@0.5, and remains closely behind the best-performing method, HEAL, at AP@0.7.
We further analyze robustness to localization noise in \cref{fig:noise}(a), where AG-PGO drives our method's resilience. Most baselines degrade sharply at the strict AP@0.7 metric; HyDRA mitigates this more effectively than most heterogeneity-aware baselines and remains comparable to GenComm across thresholds. At the low-threshold AP@0.3, HyDRA also matches MPDA's object recovery, 
though MPDA retains a consistent edge over HyDRA at the stricter AP@0.5 and AP@0.7 thresholds \red{(full results in supp.\ Sec.~F)}.

Thus, HyDRA matches the heterogeneity-aware baselines without any retraining or parameter updates, whereas these baselines require additional training or model adaptation to handle unseen agents.

\subsection{Analysis of Latent Domain Heterogeneity}
\label{subsec:latent_domain_analysis}

\red{Prior heterogeneity-aware methods align an agent to the ego domain only when its declared configuration differs; when configurations match, no alignment is triggered. This raises a fundamental question: \textit{Is relying solely on architectural metadata sufficient to identify heterogeneity?} To answer it, we compare naive fusion strategies in a setting where all agents share identical architectural metadata, so that these baselines would effectively reduce to Intermediate Fusion (\cref{tab:res}).}
Despite the matching metadata, Intermediate Fusion fails to surpass Late Fusion in overall AP across all IoU thresholds, 
\red{indicating that fusing $X_P$'s features degrades rather than complements the ego's representation. As all agents share the same declared architecture, this row serves as a proxy for routing by declared metadata alone, showing that such metadata is insufficient.}

Since our domain classifier probes feature compatibility directly from transmitted features, it can serve as a general-purpose plug-in that triggers a baseline's alignment mechanism whenever matching declared architectures would otherwise leave it dormant. We validate this by equipping each heterogeneity-aware baseline (MPDA, HEAL, CodeFilling, GenComm) with our classifier to route $X_P$ accordingly; as \cref{tab:res} shows, every classifier-equipped baseline substantially outperforms naive fusion, confirming that the classifier supplies the missing trigger signal these methods need. Among classifier-equipped methods, 
\red{HyDRA remains competitive without additional training, although CodeFilling outperforms it at AP@0.5 and AP@0.7, and MPDA at AP@0.5,} and we observe a similar trend on OPV2V (\cref{tab:opv2v_combined}, \textbf{L} columns).

\subsection{Scalability and Efficiency Analysis}
Unlike baselines that require retraining and incur parameter growth for each new heterogeneous agent---including PHCP~\cite{si2025you}, which performs inference-time adaptation via gradient-based self-training of an adapter, taking 1.5--2.4\,s per agent as reported---HyDRA scales without parameter growth, as reflected by the zero-cost metric in \cref{fig:noise}(b). 
\red{Its routing overhead is a gradient-free 9.4\,ms evaluation per agent per frame (supp.\ Sec.~D), over one to $N{=}10$ frames, versus PHCP's reported one-time cost of 1.5--2.4\,s, and} without requiring even a small labeled or pseudo-labeled support set. We compare against PHCP on latency only, due to differences in its official evaluation protocol.

\begin{figure}[t]
    \centering

    \includegraphics[width=1.0\linewidth]{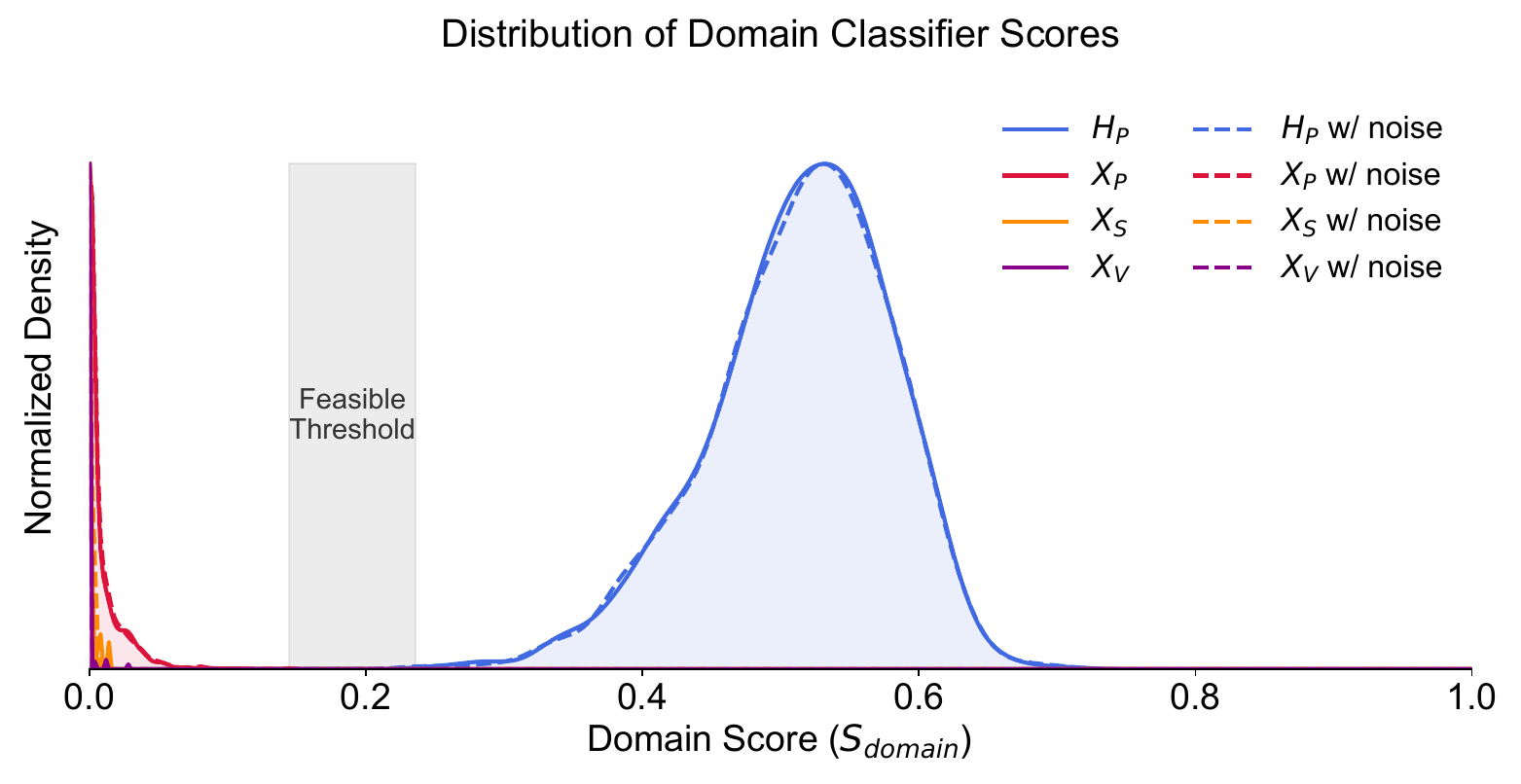}
    \caption{Analysis of domain scores comparing homogeneous vs. heterogeneous agents under ideal and noisy ($\sigma=0.4$) conditions.}
    \label{fig:domain_classifier}
\end{figure}

\subsection{Component Analysis}

\subsubsection{Domain Classifier}\label{subsec:dc_analysis}
\cref{fig:domain_classifier} shows the score distributions. Despite sharing the identical PointPillars architecture, the homogeneous agent ($H_P$) receives a high domain score while the latent heterogeneous agent ($X_P$) scores near zero—closely matching the near-zero scores of the architecturally heterogeneous agents ($X_S, X_V$)—
\red{indicating that the score reflects feature compatibility rather than declared architecture.}

\red{The score distributions are virtually unchanged under pose noise, and the minimum homogeneous score exceeds the maximum heterogeneous score, leaving a broad feasible threshold region ($[0.166, 0.236]$ on V2X-Real; supp.\ Sec.~B).}
\red{Replacing Soft-AP with mean IoU collapses this separation, }
confirming that the separation stems from our Soft-AP formulation rather than from IoU matching alone.

We verify that this separation generalizes across datasets, training checkpoints, and ego backbones. On OPV2V, the classifier maintains a $[0.031, 0.442]$ feasible region that contains V2X-Real's, so a single threshold transfers across datasets. 
\red{Across two additional random seeds (same single-agent recipe),} $X_P$'s score consistently stays well below $H_P$'s minimum ($X_P$ max 0.166 vs.\ $H_P$ min 0.246), showing that the separation is not tied to a specific checkpoint. 
Replacing the ego with SECOND also preserves a clear margin between the homogeneous and heterogeneous groups (\cref{tab:ego_generality}), suggesting that the classifier is not specific to the ego's architecture.

\begin{table}[t]
\centering
\small
\setlength{\tabcolsep}{8pt}
\begin{tabular}{cl ccc}
\toprule
\multirow{2}{*}{\textbf{Ego}} & \multirow{2}{*}{\textbf{Agent Type}} & \multicolumn{3}{c}{\textbf{Domain Score ($S_{\text{domain}}$)}} \\
\cmidrule(lr){3-5}
 & & \textbf{Mean} & \textbf{Max} & \textbf{Min} \\
\midrule
\multirow{2}{*}{$E_S$}
 & Homogeneous & 0.5747 & 0.7415 & 0.2303 \\
 & Heterogeneous & 0.0007 & 0.0698 & 0.0000 \\
\bottomrule
\end{tabular}
\caption{\red{Domain scores with SECOND as the ego ($E_S$). Homogeneous: $H_S$; heterogeneous: $X_S, X_P, X_V$.}}
\label{tab:ego_generality}
\end{table}

\begin{table}[t]
\centering
\small
\setlength{\tabcolsep}{5pt}
\begin{tabular}{cc cccccc}
\toprule
\multirow{2}{*}{\textbf{Agent}} & \multirow{2}{*}{\textbf{$N$}} & \multicolumn{6}{c}{\textbf{Shared Objects per Frame ($K$)}} \\
\cmidrule(lr){3-8}
 & & \textbf{1} & \textbf{2} & \textbf{3} & \textbf{5} & \textbf{10} & \textbf{20} \\
\midrule
\multirow{3}{*}{$H_P$}
 & 1  & 41.94 & 29.58 & 22.74 & 13.14 & 2.31 & 0.03 \\

 & 2  & 30.43 & 18.22 & 12.14 & 5.01  & 0.35 & 0.00 \\

 & 10 & 3.02  & 0.85  & 0.25  & 0.02  & 0.00 & 0.00 \\
\midrule
\multirow{3}{*}{$X_P$}
 & 1  & 2.99 & 4.66 & 5.27 & 2.87 & 0.96 & 0.22 \\

 & 2  & 4.66 & 4.24 & 1.99 & 0.99 & 0.35 & 0.08 \\

 & 10 & 0.91 & 0.21 & 0.06 & 0.01 & 0.00 & 0.00 \\
\midrule
$X_{S/V}$ & 1 & 0.07 & 0.10 & 0.09 & 0.02 & 0.00 & 0.00 \\
\bottomrule
\end{tabular}
\caption{Misclassification rate (\%) at threshold $\tau=0.135$ under sparse shared observations ($K$), before ($N{=}1$) and after accumulating domain scores over $N$ consecutive frames.}
\label{tab:subsampling}
\end{table}

\red{Since the score is computed over shared detections, we stress-test the classifier by subsampling shared objects to $K$ per frame (\cref{tab:subsampling}, $\tau{=}0.135$). At $N{=}1$, heterogeneous agents ($X_S, X_V, X_P$) are rejected with low error across all $K$, while homogeneous agents ($H_P$) are more sensitive to sparsity; these errors are conservative, merely routing $H_P$ to late fusion.} 
\red{Soft-AP also outperforms a Hard-AP variant once $K{\geq}5$ (supp.\ Sec.~E).}
\red{Accumulating scores over 10 frames drives the error toward zero for nearly all $K$.}

\subsubsection{AG-PGO}
To validate the efficacy of our anchor-guided strategy, we compare the standard optimization module (CoAlign) against our proposed AG-PGO in the noisy setting.
The results in \cref{tab:results}(b) indicate that AG-PGO consistently yields higher detection accuracy across all AP metrics compared to the standard CoAlign baseline.
\red{Since only heterogeneous agents' poses are optimized, AG-PGO also converges within 50 iterations, versus approximately 500 for CoAlign.}

\subsection{Ablation Study}

\red{We ablate each component under architecture heterogeneity with pose noise (\cref{tab:ablation_study}). Removing the domain classifier causes the largest drop; here we invert the routing as a worst case (homogeneous $\rightarrow$ late, heterogeneous $\rightarrow$ intermediate), which shows the cost of misrouting rather than comparing routing rules. For the latter, routing by declared architecture equals oracle routing in this setting, which the classifier matches as the two groups' scores are fully separated (\cref{subsec:dc_analysis}), whereas under latent heterogeneity metadata-based routing reduces to Intermediate Fusion in \cref{tab:res}. Omitting AG-PGO also degrades performance, though less severely.}

\begin{table}[t]
\centering
\resizebox{\columnwidth}{!}{%
\begin{tabular}{cc ccc}
\toprule
\textbf{Domain Classifier} & \textbf{AG-PGO} & \textbf{AP@0.3} & \textbf{AP@0.5} & \textbf{AP@0.7} \\
\midrule
-          & -          & 0.2559 & 0.1173 & 0.0106 \\
-          & \checkmark & 0.2827 & 0.1627 & 0.0245 \\
\checkmark & -          & 0.4886 & 0.3829 & 0.1568 \\
\checkmark & \checkmark & 0.5093 & 0.3904 & 0.2164 \\
\bottomrule
\end{tabular}%
}

\caption{\red{Ablation under architecture heterogeneity with pose noise ($\sigma{=}0.4$). Without the classifier, routing is inverted as a worst case (homogeneous $\rightarrow$ late, heterogeneous $\rightarrow$ intermediate).}}
\label{tab:ablation_study}
\end{table}

%% file: sec/5_conclusion.tex
\section{Conclusion and Limitations}\label{Sec:Conclusion}

\red{HyDRA probes each agent's feature compatibility at inference time and routes it to intermediate or late fusion without retraining, performing on par with heterogeneity-aware methods. Its contribution is a simple, training-free integration of a compatibility probe with existing fusion and pose-graph mechanisms rather than a new fusion operator.
Several limitations remain. The classifier assumes a common BEV feature interface, and we evaluate LiDAR agents only. Its score measures decodability, a proxy for, rather than a guarantee of, beneficial fusion, relying on a competent ego decoder and reliable sender detections. Our latent-heterogeneous agent differs from the homogeneous one in training run and regime, which we do not disentangle. AG-PGO does not correct homogeneous agents' poses, relying on feature-level fusion to partially compensate for their errors. Sparse observations defer routing; onboarding and homogeneous agents send full feature maps; the threshold region may need revisiting elsewhere, and late-fusion agents trail the most noise-robust baselines under severe noise.}